\pdfoutput=1 % make arXiv use pdfLaTeX
\documentclass{article} % For LaTeX2e
\usepackage[final]{colm2026_conference}
\usepackage[T1]{fontenc} % needed for \k{a} (e.g. M\k{a}dry) in the bibliography

\usepackage{microtype}
\usepackage{hyperref}
\usepackage{url}
\usepackage{booktabs}
\usepackage{amsmath,amssymb,amsfonts}
\usepackage{graphicx}
\usepackage{xcolor}
\usepackage{lineno}
\usepackage{wrapfig}
\usepackage{microtype}

\setcitestyle{numbers,square}

\definecolor{darkblue}{rgb}{0, 0, 0.5}
\hypersetup{colorlinks=true, citecolor=darkblue, linkcolor=darkblue, urlcolor=darkblue}

\title{Pattern-Driven Evaluation of LLM Code Optimization: \\ Isolating Compiler-Transcendent Semantic Reasoning}

\author{Wesley Lu \\
Georgia Institute of Technology \\
\texttt{wesleylu@gatech.edu}
\And
Runzhou Liu \\
University of Virginia \\
\texttt{azd9hw@virginia.edu}
\And
Hongru Du \\
University of Virginia \\
\texttt{hongrudu@virginia.edu}
\And
Xinyao Yi \\
University of Virginia \\
\texttt{mrq9gz@virginia.edu}
}

\begin{document}

\ifcolmsubmission
\linenumbers
\fi

\maketitle

\begin{abstract}
Existing evaluations of LLM-based code optimization primarily measure end-to-end speedup, obscuring whether a model actually performs the intended optimization. For compiled languages, this is especially problematic: observed gains may reflect transformations already recoverable by modern optimizing compilers rather than optimization behavior attributable to the model itself. We present \textbf{PatternBench}, a benchmark for studying compiler-transcendent optimization, i.e., performance improvements that remain after aggressive compiler optimization. PatternBench targets residual optimization opportunities across seven classes of transformations and pairs runtime evaluation with an AST-based optimization faithfulness analysis that verifies whether the intended optimization is realized in the generated program. Experiments with frontier, open-weight, and fine-tuned models show that runtime speedup alone provides an incomplete account of optimization capability. Optimization faithfulness is strongly associated with realized speedup, and fine-tuning improves both faithfulness and optimization effectiveness. These results suggest that evaluation of LLM-based code optimization should measure not only performance outcomes, but also whether the target transformation was successfully executed. 
\end{abstract}

\section{Introduction}

Large language models (LLMs) are increasingly capable of improving program performance, with recent work demonstrating measurable runtime improvements across a variety of code optimization benchmarks \cite{du2024mercury,huang2024effibench,liu2024evalperf,qiu2024enamel,he2025sweperf,shypula2023pie}. These results suggest that LLMs may become useful assistants for performance engineering and compiler-assisted software development.

Evaluating optimization capability, however, remains challenging. Unlike functional code generation, optimization success cannot be fully characterized by correctness alone. Modern compilers already perform a wide range of sophisticated optimizations, including inlining, vectorization, loop transformations, and interprocedural analysis \cite{aho2007compilers}. Consequently, performance gains reported by existing benchmarks often conflate optimizations generated by the model with transformations that are already within the capabilities of modern compiler technology. This ambiguity raises two important questions. First, can LLMs identify optimization opportunities that remain beyond aggressive compiler optimization? Second, when a model proposes an optimization strategy, does it actually realize that transformation in the generated code? Existing optimization benchmarks primarily evaluate end-to-end speedup and provide limited insight into either question.

To address these limitations, we introduce PatternBench, a benchmark designed around compiler-transcendent optimization opportunities—optimization patterns that survive aggressive compiler optimization and therefore require reasoning beyond standard compiler analyses. In addition to measuring performance improvement, PatternBench incorporates an AST-based optimization faithfulness analysis that verifies whether the intended optimization transformation is actually realized in the generated program.

Using a collection of frontier, open-weight, and fine-tuned language models, we study both optimization effectiveness and optimization faithfulness. Our results reveal substantial variation in compiler-transcendent optimization capability across models and show that optimization faithfulness is strongly associated with realized performance gains. Models that consistently implement intended transformations achieve significantly larger speedups, while models with lower faithfulness frequently fail to translate optimization plans into effective code modifications. The contributions of this work are as follows:
\begin{itemize}
    \item A benchmark of compiler-transcendent optimization tasks designed to isolate optimization opportunities that survive aggressive compiler optimization.
    \item An AST-level verification methodology that enables process-aware evaluation of optimization behavior through optimization faithfulness.
    \item A comparative study of frontier, open-weight, and fine-tuned language models on compiler-transcendent optimization.
    \item Empirical evidence that optimization faithfulness predicts realized performance gains and exposes optimization failures missed by speedup-only evaluation.
\end{itemize}

\section{Related Work}

\paragraph{LLM-Based Code Optimization.} Recent work has demonstrated that large language models can improve program performance through source-level code transformations. Mercury~\cite{du2024mercury}, EffiBench~\cite{huang2024effibench}, EvalPerf~\cite{liu2024evalperf}, ENAMEL~\cite{qiu2024enamel}, PIE~\cite{shypula2023pie}, and SWE-Perf~\cite{he2025sweperf} evaluate optimization capability through runtime improvement relative to human-written implementations or expert-generated patches. Complementary methods improve efficiency at inference time by feeding execution logs back to the model for iterative self-optimization~\cite{huang2025effilearnerenhancingefficiencygenerated}. These benchmarks have significantly advanced the study of LLM-driven optimization.  However, evaluation is primarily outcome-based: models are rewarded for achieving speedups, regardless of whether the intended optimization was actually realized in the generated program. In contrast, PatternBench complements performance evaluation with structural verification of optimization execution.

\paragraph{Compiler Optimization and Superoptimization.}
Modern compilers implement a wide range of optimizations, including inlining, vectorization, loop transformations, interprocedural optimization, and profile-guided optimization~\cite{aho2007compilers}. Learning-based compiler systems such as MLGO~\cite{trofin2021mlgo} and ProGraML~\cite{cummins2021programl} improve optimization decisions within existing compiler frameworks, while superoptimization systems including STOKE~\cite{schkufza2013stoke}, Souper~\cite{sasnauskas2017souper}, and AlphaDev~\cite{mankowitz2023alphadev} search for provably better implementations at the instruction level. More recently, LLMs have been applied directly to compiler optimization~\cite{cummins2023largelanguagemodelscompiler,cummins2024metalargelanguagemodel}, including agentic systems that couple compiler passes with LLM-generated edits~\cite{mikek2026agenticcodeoptimizationcompilerllm}. Although highly effective, these approaches largely operate within a semantics-preserving optimization space defined by compiler analyses and transformation rules. PatternBench focuses instead on residual optimization opportunities that survive aggressive compiler optimization and therefore require reasoning beyond conventional compiler analyses.

\paragraph{Optimization Verification and Faithfulness.}
Prior work has shown that model-generated explanations do not always faithfully reflect the factors driving model behavior~\cite{turpin2023language,lanham2023measuring,chen2025reasoningmodels}. Recent work frames the reasoning trace as a fragile but valuable channel for monitoring model behavior~\cite{korbak2025chainthoughtmonitorabilitynew} and builds dedicated tests for explanation faithfulness~\cite{matton2025walktalkmeasuringfaithfulness,shen2025faithcotbench}. While these studies focus primarily on natural-language reasoning, they highlight a broader challenge of determining whether a claimed reasoning process is actually reflected in system behavior. In software engineering, transformation validation and optimization verification techniques have been developed to verify correctness and equivalence of compiler transformations~\cite{lecong2023invalidator,wu2025fse}. Rather than evaluating optimization solely through runtime, we introduce an AST-based optimization faithfulness analysis that verifies whether the intended optimization transformation is realized in the generated program, enabling process-aware evaluation of LLM-based code optimization.

\section{PatternBench}

PatternBench is designed to evaluate compiler-transcendent optimization opportunities, i.e., performance improvements that remain available after aggressive compiler optimization. The benchmark is constructed around two requirements. First, benchmark instances must contain optimization opportunities that are not recovered automatically by the compiler configurations considered in this study. Second, the target optimization must be structurally verifiable, enabling downstream analysis of optimization faithfulness.

\begin{figure}
    \centering
    \includegraphics[width=1\linewidth]{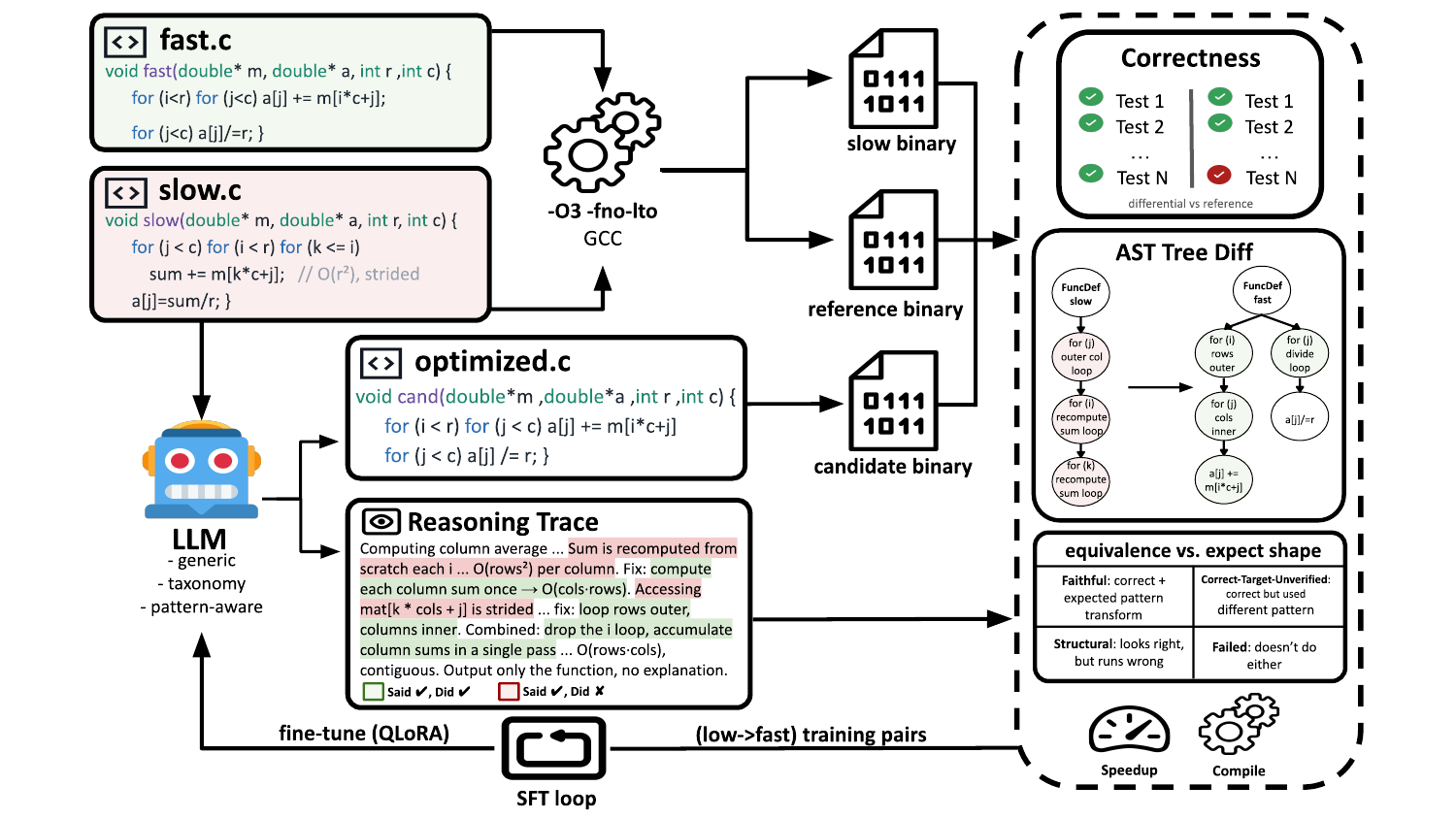}
    \caption{Overview of PatternBench. Each instance pairs a slow baseline with a hand-written reference, both of which are compiled with GCC's highest level of aggressive, file-specific optimization to ensure the slow/fast gap survives compiler optimization. An LLM or QLoRA-tuned model emits the optimized code, which is evaluated for correctness, compilation, speedup, and faithfulness to the target transformation. Reasoning traces are recorded for analysis.}
    \label{fig:placeholder}
\vspace{-5mm}
\end{figure}
\vspace{-2mm}

\subsection{Compiler-Transcendent Optimization}

We define a compiler-transcendent optimization (CTO) as an optimization whose performance benefit is not recovered by the compiler configurations considered in this study. Such opportunities arise when profitable transformations depend on information that is unavailable to compiler analyses or cannot be exploited safely within conventional optimization frameworks. Typical examples include optimizations requiring semantic interpretation of program intent, runtime-dependent specialization, algorithmic substitution, or representation-level redesign. 
% The purpose of this definition is operational rather than theoretical. 
CTOs provide a mechanism for separating optimization opportunities that are already handled by existing compiler pipelines from those that potentially require additional reasoning capabilities.

\subsection{Benchmark Construction}

PatternBench is organized around seven classes of residual optimization opportunities observed during preliminary analysis of compiler-resistant performance bottlenecks. The categories span semantic simplification, control-flow restructuring, algorithmic improvement, data-structure transformation, runtime specialization, and compositional optimizations requiring multiple coordinated transformations.

Each benchmark instance contains a baseline implementation, an optimized reference implementation, and an associated optimization pattern used for downstream verification. Table~\ref{PatternBench_composition} summarizes the benchmark composition.

\subsection{Compiler-Resistance Validation}

A benchmark intended to evaluate compiler-transcendent optimization must establish that the optimization opportunity remains available after compiler optimization. Otherwise, measured improvements cannot be attributed to capabilities beyond the compiler itself.

To this end, both the baseline and the optimized reference are compiled with GCC at \texttt{-O2 -fno-lto} (link-time optimization disabled so that single-translation-unit resistance is not masked by whole-program inlining). Instances are retained only when a measurable performance gap persists between the baseline and optimized implementations. Cases in which the compiler can automatically eliminate the gap are excluded from the benchmark.

This validation procedure ensures that benchmark instances correspond to residual optimization opportunities rather than missed compiler optimizations. As a result, performance improvements on PatternBench can be interpreted as evidence of optimization behavior beyond the transformations captured by the compiler configurations evaluated.

\vspace{-3mm}
\begin{table}[!htbp]
\centering
\small
\begin{tabular}{lcc}
\toprule
Category & Total & Description \\
\midrule
Semantic Redundancy (SR) & 112 & Remove redundant computation the compiler cannot hoist \\
Input-Sensitive (IS)     & 96 & Specialize for runtime input distributions \\
Algorithmic (AL)         & 78 & Replace with lower-complexity algorithms \\
Memory/IO (MI)           & 77 & Improve memory access patterns and locality \\
Data-Structure (DS)      & 76 & Change data representation or layout \\
Human-Style (HR)         & 75 & Recover a fast form obscured by code structure \\
Control-Flow (CF)        & 55 & Restructure non-pure control flow \\
Compositional (COMP)     & 675 & Multiple coordinated transformations \\
\midrule
Total & 1244 & \\
\bottomrule
\end{tabular}
\caption{PatternBench composition. Numbers represent combined base single-family and compositional patterns.}
\label{PatternBench_composition}
\end{table}

\section{Optimization Faithfulness}\label{sec:faithfulness}

Performance improvement alone does not fully characterize optimization behavior. A generated program may achieve a speedup without implementing the optimization strategy identified by the model, while a model may correctly recognize an optimization opportunity but fail to realize it in code. In both cases, runtime measurements provide limited visibility into how the optimization was achieved.

This issue is particularly relevant for LLM-based optimization because optimization is fundamentally a transformation task. Beyond determining whether the resulting program is faster, it is often desirable to establish whether the intended transformation was actually carried out. PatternBench therefore evaluates optimization behavior along two dimensions: optimization effectiveness, measured through runtime improvement, and optimization faithfulness, measured through structural verification of the target transformation.

\subsection{Faithfulness Verification}

Each PatternBench instance is associated with a target optimization pattern describing the intended transformation. For example, an instance may require elimination of redundant computation, loop-invariant code motion, data-structure replacement, or control-flow simplification. We score a model's emitted code on two independent axes and cross them into a single verdict, rather than trusting the model's account of what it did, since chain-of-thought is an unreliable guide to the underlying computation~\cite{turpin2023language,lanham2023measuring,chen2025reasoningmodels}.

We classify outputs as Faithful, Correct--Target-Unverified, Structural-Only, or Failed. Because Correct--Target-Unverified absorbs checker misses, the Faithful rate is a lower bound on how often a model applies a valid optimization. We discuss verifier validation as a limitation (\S\ref{sec:limitations}). This limitation does not reduce the procedure to outcome-based evaluation. Unlike a speedup measurement, our verifier examines whether the optimization itself appears in the emitted code. A speedup alone is insufficient evidence of successful optimization because performance improvements may arise from unrelated modifications, compiler effects, or measurement noise.

\subsection{Faithfulness Metric}

For each model attempt $a$, let $C(a) = 1$ if the generated program is behaviorally correct and $S(a) = 1$ if the AST verifier detects the target optimization pattern. We define $F(a) = C(a)S(a)$. For model m with attempts $A_m$, faithfulness is $Faithfulness(m) = \sum_{a\in A_m}\frac{F(a)}{|A_m|}$. Faithfulness is reported both at the benchmark level and for individual optimization categories. 
% This enables analysis of which classes of optimization opportunities are consistently realized by a model and which remain difficult to execute. 
Attempts with $C(a)=1$ and $S(a)=0$ form the Correct--Target-Unverified category. The program passes differential testing, but the checker does not perform the target transformation. 
We report it alongside speedup because speedup measures performance improvement, whereas faithfulness measures optimization execution. 

% Faithfulness is the fraction of all attempts in which the emitted code is both behaviorally equivalent to the reference and structurally realizes the intended transformation: $Faithfulness = Faithful Solutions / All Attempts$.

% \[
% \text{Faithfulness} =
% \frac{\#\,\text{Faithful}}
% {\#\,\text{All Attempts}}
% \]

\section{Evaluation}

\subsection{Experimental Setup}

We evaluate each model on PatternBench under three prompting conditions that vary the amount of optimization guidance provided to the model. The \emph{generic} prompt gives only the source function and asks the model to optimize it. The \emph{taxonomy-guided} prompt additionally provides the seven optimization categories used by PatternBench. The \emph{pattern-aware} prompt provides the exact optimization category and pattern description for the instance. These settings separate failures of recognition from failures of implementation: the generic setting requires the model to identify the optimization opportunity, whereas the pattern-aware setting primarily tests whether the model can carry out a known transformation. All prompts require the model to return a portable C implementation of a function named \texttt{optimized}; architecture-specific intrinsics such as SSE, AVX, or AVX2 are disallowed.

We also evaluate fine-tuned open-weight models to study whether compiler-transcendent optimization patterns can be learned from examples. Fine-tuning is performed on paired slow and optimized implementations drawn from the training split. Held-out instances are separated at the variant level, so that a held-out instance does not appear in training under any prompting condition. We report results separately for base and fine-tuned models.

For each generated program, we measure functional correctness, runtime speedup, and optimization faithfulness. Correctness is checked with differential testing across nine input configurations and data type specific tolerances. Runtime is measured after compilation under the same compiler configuration used for evaluation. Optimization faithfulness is computed using the AST-based verifier described in Section~\ref{sec:faithfulness}.

\subsection{Metrics}

We evaluate model performance along three dimensions: correctness, optimization effectiveness, and optimization faithfulness. 1) A generated solution is considered correct if it passes all benchmark test cases. Since performance improvements are meaningful only when program semantics are preserved, all optimization analyses are restricted to functionally correct solutions. 2) Optimization effectiveness is measured as runtime speedup relative to the baseline implementation. For a benchmark instance with baseline runtime $T_b$ and optimized runtime $T_o$, speedup is defined as $T_b / T_o$. Following prior optimization benchmarks~\cite{du2024mercury,huang2024effibench,liu2024evalperf,qiu2024enamel,he2025sweperf,shypula2023pie}, we report the geometric mean speedup across benchmark instances. 3) Optimization faithfulness is measured using the AST-based verification framework described in Section~\ref{sec:faithfulness}. We report benchmark-level faithfulness as well as category-level faithfulness when analyzing model behavior across optimization classes.

% \paragraph{Correctness.}
% A generated solution is considered correct if it passes all benchmark test cases. Because performance improvements are meaningful only when program semantics are preserved, speedup is reported only over functionally correct solutions.

% \paragraph{Optimization Effectiveness.}
% Optimization effectiveness is measured as runtime speedup relative to the baseline implementation. For a benchmark instance with baseline runtime $T_b$ and optimized runtime $T_o$, speedup is computed as $\text{Speedup} = T_b / T_o$. Following prior optimization benchmarks~\cite{du2024mercury,huang2024effibench,liu2024evalperf,qiu2024enamel,he2025sweperf,shypula2023pie}, we report geometric-mean speedup across benchmark instances.

% \paragraph{Optimization Faithfulness.}
% Optimization faithfulness is measured using the AST-based verification framework described in Section~\ref{sec:faithfulness}. We report benchmark-level faithfulness as well as category-level faithfulness when analyzing model behavior across different optimization classes.

\section{Findings}

\subsection{Optimization Faithfulness Predicts Optimization Effectiveness}
Optimization faithfulness and realized speedup are strongly associated (Spearman $\rho = 0.83$ across the full benchmark).
Complete benchmark evaluations are presented in Appendix~\ref{app:results}.
Fig.~\ref{fig:figure2} plots the two on the in-distribution held-out set: models with higher faithfulness consistently achieve larger performance gains, whereas models with low faithfulness remain confined to substantially smaller speedups. This pattern is visible both across external baselines and within our own model family. Weak base models such as \texttt{deepseek-r1-distill-qwen-1.5b} and \texttt{qwen2.5-coder-1.5b} lie in the lower-left region of the plot, while the upper-right region contains high-performing models such as \texttt{claude-sonnet-4-6}, \texttt{r1-32b-FT}, and \texttt{qwen-coder-32b-FT}.

\begin{figure}
      \centering
      \includegraphics[width=.8\linewidth]{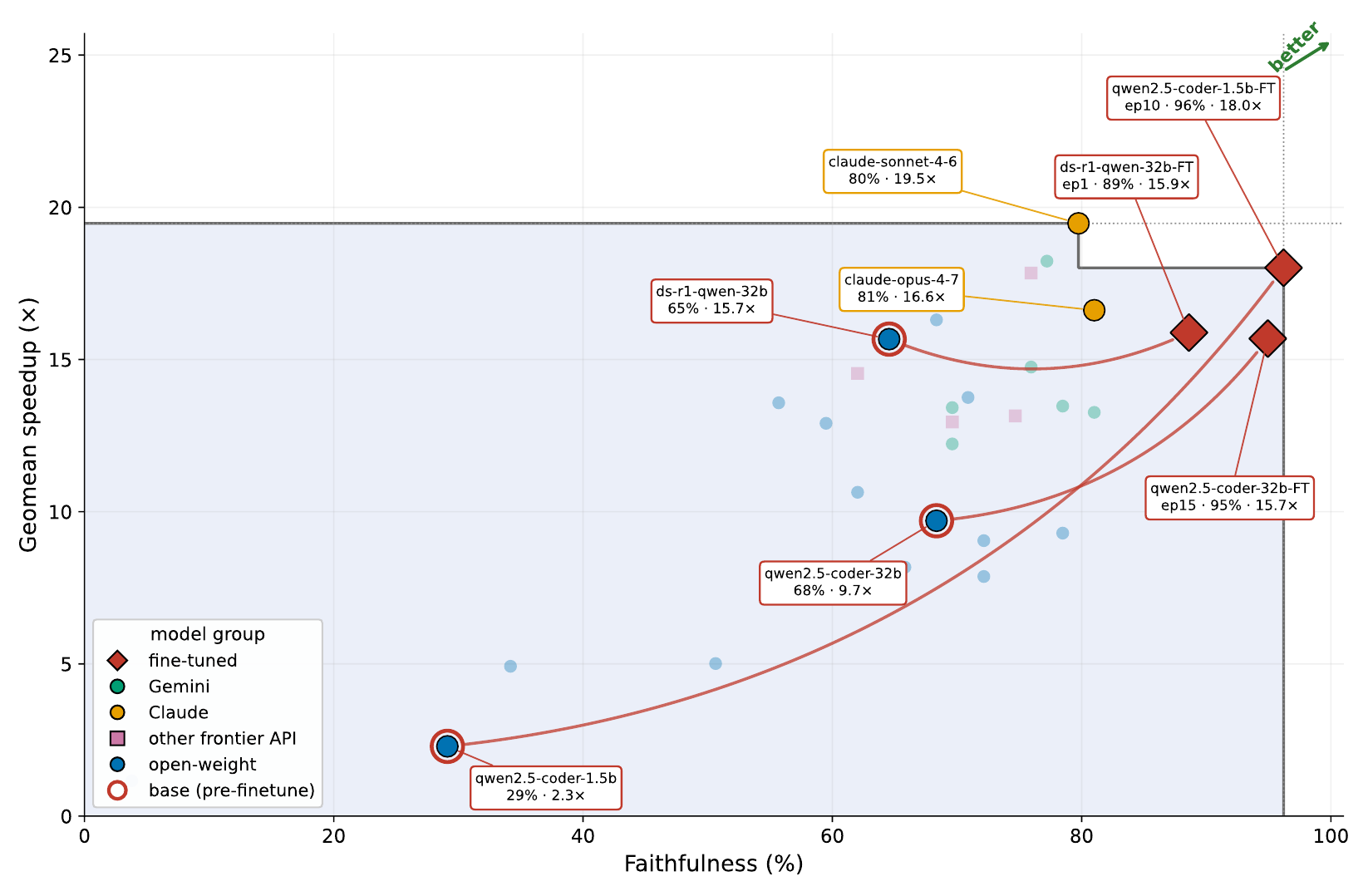}
      \vspace{-5mm}
      \caption{Optimization faithfulness versus geometric-mean speedup on the in-distribution held-out set.
      Each point denotes one model. Fine-tuned models move consistently toward the upper-right region, indicating that higher faithfulness is associated with larger realized speedup.
      }
      \vspace{-5mm}
    \label{fig:figure2}
\end{figure}

Runtime speedup alone cannot show whether a model performed the intended optimization, since faster code may result from incidental changes. Optimization faithfulness addresses this by checking whether the expected transformation is structurally present. The pattern in Fig.~\ref{fig:figure2} suggests that PatternBench speedups are largely driven by faithful execution of the intended transformation rather than unrelated edits.

This trend is especially clear after fine-tuning, as the models consistently move toward higher faithfulness and, for lower-speedup bases, higher realized speedup. Overall, performance gains on compiler-transcendent optimization tasks depend not only on recognizing profitable opportunities but also on reliably implementing the correct transformations.

\subsection{Fine-Tuning Improves Optimization Execution}

The fine-tuned models in Figure~\ref{fig:figure2} shift upward in optimization faithfulness and, for models whose base speedup is not already saturated, rightward in realized speedup. For \texttt{qwen2.5-coder-1.5b}, fine-tuning improves faithfulness from 29\% to 96\% and speedup from 2.3$\times$ to 18.0$\times$; for \texttt{qwen2.5-coder-32b}, it improves faithfulness from 68\% to 95\% and speedup from 9.7$\times$ to 15.7$\times$; and for \texttt{ds-r1-qwen-32b}, it improves faithfulness from 65\% to 89\%, while speedup is essentially unchanged. Across the three models, faithfulness rises by 24 to 67 percentage points with the largest gain where base faithfulness was lowest.

The consistency of this improvement indicates that fine-tuning closes an execution gap, improving the model's ability to carry out target transformations rather than producing superficially optimized code, though this converts into additional speedup only where the base model has not already exhausted the available headroom. The effect is largest for the smallest model, \texttt{qwen2.5-coder-1.5b}, suggesting that much of the gap between weak and strong open-weight models on PatternBench lies in transformation execution rather than optimization recognition.

\subsection{Prompt Guidance Separates Recognition from Execution Failures}\label{subsec:prompt-guidance}

A model's faithfulness score conflates two abilities: recognizing which optimization is needed and executing it correctly. Our three prompting conditions separate them by varying the recognition burden: the generic prompt supplies only the source function, the taxonomy-guided prompt narrows the choice to the seven categories, and the pattern-aware prompt names the exact target so that failures chiefly reflect execution rather than recognition. Across the 22 models evaluated under all three strategies, faithfulness rises from 35.5\% (generic) to 45.6\% (pattern-aware), an average gain of 10.1 percentage points. A large generic-to-pattern-aware gain marks a model as recognition-limited, able to perform the rewrite once it is named but unable to infer it from code alone; the largest gains occur for qwen2.5-coder-14b (+20), qwen2.5-coder-7b (+17), glm-5 (+16), claude-opus (+15), and qwen3-coder-30b (+14), whereas a small gap points to an execution bottleneck. That execution bottleneck is substantial: even under pattern-aware prompting, mean faithfulness reaches only 45.6\% (maximum 70.0\%), so over half of attempts still fail to realize the named transformation.

% A speedup or faithfulness score conflates abilities in recognizing the needed optimization and executing it correctly. Our three prompting conditions that vary the recongition burden. The prompt requires both recognition and execution. The generic prompt requires both recognition and execution. The pattern-aware prompt gives the exact target pattern and isolating execution. The taxonomy-guided prompt is intermediate, narrowing recognition to seven choices. Thus, generic faithfulness measures end-to-end performance, pattern-aware faithfulness measures execution given the correct diagnosis, and their gap estimates the recognition bottleneck. 

% Moving from generic to pattern-aware prompting raises faithfulness by $\text{FILL IN}$ percentages on average between all evaluated models, from $\text{FILL IN}$ to $\text{FILL IN}$. The largest gains occurs on $\text{FILL IN}$, suggesting that these models can often perform the target rewrite once it's named, but fails to identify it unprompted. Models with small gaps are instead limited mainly by execution. Even under the pattern-aware condition, faithfulness remains well below what we expected. This is because the model was handed the exact diagnosis and still doesn't realize the transformation in code. They concentrate in structural and algorithmic categories where the transformation requires coordinated non-local rewriting rather than localized edits. 

\subsection{Compiler-Transcendent Optimization Remains Challenging for Structural and Algorithmic Transformations}

\begin{wraptable}{r}{0.7\textwidth}\small
\vspace{-0.7em}
  \begin{tabular}{lcccccccr}
    \toprule
    Model & SR & IS & CF & DS & AL & MI & HR & COMP \\
    \midrule
    claude-opus-4-7 & 90 & 71 & 69 & 51 & 69 & 62 & 87 & 68 \\
    gemini-3.5-flash & 83 & 74 & 71 & 46 & 55 & 62 & 88 & 65 \\
    claude-sonnet-4-6 & 86 & 74 & 75 & 51 & 68 & 61 & 84 & 60 \\
    minimax-m3 & 82 & 75 & 71 & 41 & 72 & 62 & 73 & 58 \\
    glm-5 & 86 & 72 & 85 & 41 & 67 & 62 & 88 & 53 \\
    qwen3-32b & 51 & 72 & 71 & 37 & 72 & 62 & 69 & 53 \\
    \midrule
    \textit{all-model avg} & 62 & 62 & 59 & \textbf{34} & 50 & 60 & 70 & \textbf{42} \\
    \bottomrule
  \end{tabular}
  \vspace{-3mm}
  \caption{Per-category faithfulness (\% of all attempts, pattern-aware).
  SR--HR are single-family variants; COMP are compositional variants requiring multiple coordinated transformations. Data-structure (DS, 34\%) and compositional (COMP, 42\%) are the hardest categories.}
  \label{tab:category}
\end{wraptable}

Performance on PatternBench is not uniform across optimization categories. 
% While some residual optimizations can often be realized once the relevant pattern is recognized, others remain difficult even for strong models and under higher-guidance prompting conditions.
Some residual optimizations are realized relatively reliably, whereas others remain difficult even for strong models and under higher-guidance prompting conditions.
In our experiments, the most persistent failures arise in categories that require non-local program restructuring, algorithmic substitution, or coordinated changes to data representation and control flow, rather than localized code edits.
\vspace{-0.7em}

% \vspace{-2mm}

% \begin{table}[!htbp]
%   \centering
%   \small
%   \setlength{\tabcolsep}{6pt}
%   \begin{tabular}{lcccccccr}
%     \toprule
%     Model & SR & IS & CF & DS & AL & MI & HR & COMP \\
%     \midrule
%     claude-opus-4-7 & 90 & 71 & 69 & 51 & 69 & 62 & 87 & 68 \\
%     gemini-3.5-flash & 83 & 74 & 71 & 46 & 55 & 62 & 88 & 65 \\
%     claude-sonnet-4-6 & 86 & 74 & 75 & 51 & 68 & 61 & 84 & 60 \\
%     minimax-m3 & 82 & 75 & 71 & 41 & 72 & 62 & 73 & 58 \\
%     glm-5 & 86 & 72 & 85 & 41 & 67 & 62 & 88 & 53 \\
%     qwen3-32b & 51 & 72 & 71 & 37 & 72 & 62 & 69 & 53 \\
%     \midrule
%     \textit{all-model avg} & 62 & 62 & 59 & \textbf{34} & 50 & 60 & 70 & \textbf{42} \\
%     \bottomrule
%   \end{tabular}
%   \caption{Per-category faithfulness (\% of all attempts, pattern-aware).
%   SR--HR are single-family variants; COMP are compositional variants requiring multiple coordinated transformations. Data-structure (DS, 34\%) and compositional (COMP, 42\%) are the hardest categories.}
%   \label{tab:category}
% \end{table}

This distinction is consistent with the nature of compiler-transcendent optimization. Transformations such as semantic simplification or elimination of redundant computation can often be implemented through relatively local edits once the inefficiency has been identified. By contrast, structural and algorithmic optimizations frequently require the model to rewrite larger portions of the program while preserving semantics, update multiple dependent program components consistently, and maintain invariants that are only implicit in the original implementation. Data-structure transformations are particularly demanding in this respect: replacing one representation with another often requires synchronized changes to allocation, access patterns, loop structure, and auxiliary state. Algorithmic substitutions impose a similar burden, since they require the model not only to recognize that a better algorithm exists, but also to re-express the computation in a semantically equivalent form.

% Together, these results indicate that the main difficulty in compiler-transcendent optimization is not recognizing profitable opportunities but reliably executing transformations whose correctness depends on coordinated structural change. 
Together, these results suggest that a substantial portion of the remaining difficulty in compiler-transcendent optimization lies in reliably executing transformations whose correctness depends on coordinated structural change.
This is also why optimization faithfulness matters: 
% failures on the harder categories arise not because the objective is unclear, but because the transformation is only partially or inconsistently implemented. 
failures on the harder categories often take the form of partial or inconsistent implementation of the intended transformation, which is not visible from speedup alone.
The category-level variation further shows that compiler-transcendent optimization is not a single scalar capability; the benchmark spans transformations that differ substantially in the semantic reasoning, structural editing, and global coordination they demand. Assessing a model therefore requires looking beyond aggregate speedup to ask which classes of transformations it can execute reliably and which remain bottlenecks.

\subsection{Chain-of-Thought as Data}
Reasoning models emit a separate reasoning trace during inference, which our harness records but never uses as a verdict input. From it we measure \emph{claim-without-doing} (CWD): how often a model commits to the expected optimization in its trace yet fails to implement it in code. A naive keyword match is biased, since it conflates merely considering a transform with committing to one and is inflated by draft code in the trace; we instead strip draft code, type the stated intent so that only an explicit commitment counts, and score the implemented axis with two independent oracles (an AST checker and an LLM judge). Across reasoning models, commitment is near-universal ($89.5$--$96.3\%$) while CWD spans $7$--$39\%$ depending on the oracle, and the two oracles agree only weakly ($\kappa = 0.04$--$0.25$), so we treat CWD as an exploratory range rather than a point estimate. Per-model rates, clustered-bootstrap confidence intervals, and the excluded truncated traces are reported in Table~\ref{tab:cwd-mvu}; tighter decoding of plan faithfulness, including a causal plan-perturbation test for load-bearingness and human-adjudicated checker validation, is left to future work.

\section{Limitations}\label{sec:limitations}

Optimization faithfulness is determined by a deterministic per-pattern checker. Because the Correct--Target-Unverified outcome absorbs correct solutions the checker does not recognize,  reported faithfulness is a lower bound rather than an exact estimate. This is a shared property of practical verification rather than a quirk of our design. Test-based correctness checks admit false-negative solutions and equivalence checks are sound only with respect to the input they exercise~\cite{qi2015plausibility,smith2015overfitting,le2014emi,lecong2023invalidator}. Our structural checker inherits the same kind of limitations, and we treat its verdicts as a lower bound accordingly. In future work, we plan to convert this into a more calibrated rate with distinct measurements on false positives, false negatives, and precision/recall. 

Compiler resistance, admission, and runtime measurements are all empirical. A variant that is compiler-resistant on this toolchain may not be on another, and the reported speedups are architecture-specific. Our claims are therefore about residual optimization under the tested regime, not hardware-independent guarantees. Code benchmarks are also susceptible to pretraining contamination~\cite{riddell2024quantifyingcontaminationevaluatingcode}. We mitigate this with the post-cutoff held-out patterns but cannot fully rule out exposure for the main set. Finally, fine-tuning gains in faithfulness concentrate on in-distribution patterns, so our fine-tuning results should be read as in-distribution improvements rather than a demonstration of general optimization skill. This is consistent with broader evidence that fine-tuning improves in-distribution performance while distorting pretrained features and weakening out-of-distribution generalization~\cite{kumar2022finetuning,chu2025sftrl}, an effect documented for low-rank adaptation and code fine-tuning in particular~\cite{biderman2024lora,wang2025codememorization}.

\section{Conclusion}

PatternBench evaluates code optimization by combining runtime speedup with a structural check that the intended transformation was actually performed. This reveals that faster code often does not reflect faithful optimization. Across frontier, open weight, and fine tuned models faithfulness, not raw speedup, better tracks true optimization ability. Fine tuning substantially improves faithful execution and can translate remaining headroom into speedup. The hardest cases involve data structure, algorithmic, and compositional rewrites, which require coordinated nonlocal changes. Overall, compiler transcendent optimization is not a single ability but a spectrum of transformation skills, and evaluating the optimization process is essential for identifying where models still fall short.

\bibliography{references}

\bibliographystyle{colm2026_conference}
\appendix

\section{Detailed Benchmark Results}
\label{app:results}

\renewcommand{\thetable}{\thesection.\arabic{table}}  % tables become A.1, A.2, ...
\setcounter{table}{0}                                  % restart counting at 1

Table~\ref{tab:main} reports performance across all 29 models on PatternBench, sorted by faithfulness. The strongest results cluster among the frontier models: \texttt{claude-opus-4-7} leads on faithfulness (70\%) while matching the top equivalence rate (90\%), with \texttt{gemini-3.5-flash} and \texttt{claude-sonnet-4-6} close behind at 67\% and 65\%. Faithfulness degrades sharply for the smallest models: the distilled 1.5B variants (\texttt{deepseek-r1-distill-qwen-1.5b}, \texttt{qwen2.5-coder-1.5b}) collapse to near-chance levels, indicating that reliably realizing the intended transform is challenging at small scale. Notably, equivalence and faithfulness are not tightly coupled: several models (e.g., \texttt{qwen2.5-coder-14b} at 86\%/43\%) produce outputs that are frequently correct on the reference inputs yet fail to capture the intended transform, underscoring the value of the pattern-aware, differential evaluation. Speedup is largely orthogonal to correctness, ranging from $1.2\times$ to over $20\times$ and driven more by code-generation style than by faithfulness. Finally, the Expert column reveals the largest remaining gap: even the best models match or beat the hand-written reference on only $\sim$30--54\% of their correct solutions, leaving substantial headroom for runtime-level optimization.

\begin{table}[!htbp]\centering\small\setlength{\tabcolsep}{6pt}
\begin{tabular}{lccccc}
\toprule
Model & Equivalent (\%) & Faithful (\%) & Speedup ($\times$) & Expert (\%) \\
\midrule
claude-opus-4-7 & 90 & 70 & 17.2 & 34 \\
gemini-3.5-flash & 90 & 67 & 20.3 & 46 \\
claude-sonnet-4-6 & 90 & 65 & 17.6 & 31 \\
gemini-3-pro & 86 & 63 & 20.2 & 38 \\
minimax-m3 & 85 & 63 & 17.9 & 30 \\
glm-5 & 84 & 62 & 18.0 & 38 \\
gemini-2.5-pro & 84 & 61 & 19.0 & 30 \\
gemini-3-flash & 84 & 60 & 20.0 & 54 \\
deepseek-v4-pro & 81 & 59 & 15.3 & 49 \\
kimi-k2.6 & 81 & 59 & 17.4 & 54 \\
qwen3-32b & 82 & 57 & 15.7 & 30 \\
qwq-32b & 77 & 57 & 14.9 & 47 \\
gemini-2.5-flash & 86 & 56 & 14.3 & 50 \\
qwen2.5-coder-32b & 79 & 55 & \phantom{0}9.3 & 32 \\
gpt-5.4-mini & 85 & 55 & 15.2 & 28 \\
qwen3-coder-30b & 74 & 50 & 13.8 & 32 \\
gemini-2.5-flash-lite & 77 & 50 & 16.2 & 19 \\
deepseek-r1-distill-qwen-32b & 70 & 47 & 16.0 & 23 \\
qwen2.5-72b & 82 & 46 & 14.1 & 31 \\
deepseek-r1-distill-llama-70b & 66 & 45 & 18.1 & 32 \\
qwen2.5-coder-7b & 81 & 44 & \phantom{0}3.8 & 12 \\
qwen2.5-coder-14b & 86 & 43 & \phantom{0}7.8 & 26 \\
codestral-22b & 66 & 42 & 13.1 & 30 \\
yi-coder-9b & 69 & 36 & 10.3 & 34 \\
opencoder-8b & 79 & 33 & \phantom{0}5.1 & 21 \\
deepseek-coder-v2-lite & 80 & 29 & \phantom{0}4.0 & 22 \\
qwen2.5-coder-1.5b & 68 & 28 & \phantom{0}1.2 & \phantom{0}8 \\
deepseek-r1-distill-qwen-7b & 31 & 17 & \phantom{0}9.8 & 22 \\
deepseek-r1-distill-qwen-1.5b & \phantom{0} 4 & \phantom{0}2 & \phantom{0}2.1 & 10 \\
\bottomrule
\end{tabular}
\caption{Benchmark model results on PatternBench (pattern-aware, all 1244 variants), sorted by faithfulness. \textbf{Equivalent}: outputs matching the reference across all nine differential input configurations. \textbf{Faithful}: fraction of all attempts realizing the intended transform. \textbf{Speedup}: geometric mean over the \texttt{-O2 -fno-lto} baseline. \textbf{Expert}: fraction of correct solutions whose runtime matches or beats the hand-written reference (\texttt{fast.c}). Even the strongest models reach the expert on only $\sim$30--54\% of cases, exposing substantial residual headroom.}
\label{tab:main}
\end{table}

Table~\ref{tab:cwd-mvu} examines a distinct failure mode we term \emph{claim-without-doing} (CWD): a model commits to an optimization in its prose plan but never implements it in the final code. Commitment is near-universal---every model articulates an optimization plan in over 89\% of cases---yet a substantial fraction of those commitments are never realized, with CWD rates of 24--39\% under the strict AST checker and 7--28\% under the LLM oracle. The low agreement between the two detectors ($\kappa$ ranging from 0.04 to 0.25) reflects how difficult it is to verify whether a stated transform was actually applied, and suggests that prose plans are a poor proxy for executed behavior.

\begin{table}[!htbp]\centering\small\setlength{\tabcolsep}{5pt}
\begin{tabular}{lccccc}
\toprule
Model & N & committed\% & CWD (strict/AST) & CWD (LLM proxy) & $\kappa$ \\
\midrule
gemini-2.5-pro        & 548 & 96.3 & 32.5\% & 23.5\% & 0.04 \\
gemini-2.5-flash      & 545 & 95.8 & 33.9\% & 21.1\% & 0.07 \\
gemini-3-flash        & 425 & 95.1 & 36.5\% & 27.8\% & 0.11 \\
gemini-3.5-flash      & 512 & 94.5 & 28.5\% & 24.6\% & 0.11 \\
gemini-3.1-pro        & 509 & 93.2 & 26.9\% & 23.4\% & 0.17 \\
r1-distill-llama-70b  & 502 & 91.4 & 34.9\% & 17.5\% & 0.25 \\
qwen3-32b             & 511 & 89.8 & 38.9\% & 18.0\% & 0.18 \\
r1-distill-qwen-32b   & 418 & 89.7 & 26.1\% & 7.4\%  & 0.13 \\
qwq-32b               & 475 & 89.5 & 24.4\% & 23.8\% & 0.08 \\
\bottomrule
\end{tabular}
\caption{\textbf{Claim-without-doing (CWD)}: models commit to optimizations but often fail to implement them. Committed\% measures how often judges detect commitment in prose plans; CWD measures the share where final code doesn't apply the transform, scored by an AST checker and LLM oracle. $\kappa$ ranges from 0.04 to 0.25, indicating substantial disagreement.}
\label{tab:cwd-mvu}
\end{table}

\end{document}